\documentclass[11pt,letterpaper]{article}
\usepackage{emnlp2017}
\usepackage{times}
\usepackage{amsfonts}
\usepackage{latexsym}
\usepackage{graphicx}

\emnlpfinalcopy

\def\emnlppaperid{***}

\title{Improving Context Aware Language Models}

\author{Aaron Jaech and Mari Ostendorf \\
  {\tt \{ajaech, ostendor\}@uw.edu}}

\date{}

\begin{document}

\maketitle

\begin{abstract}
  Increased adaptability of RNN language models leads to improved predictions that benefit many applications. 
  However, current methods do not take full advantage of the RNN structure.
  We show that the most widely-used approach to adaptation (concatenating the context with
  the word embedding at the input to the recurrent layer) is outperformed by 
  a model that has some low-cost improvements: adaptation of both the hidden and output layers. and a feature hashing bias term to capture context 
  idiosyncrasies. Experiments on language
  modeling and classification tasks using three different corpora demonstrate the advantages
  of the proposed techniques.
\end{abstract}

\section{Introduction}

The dominant paradigm for language model adaptation relies on the notion of a 
domain. Domains are in many ways inadequate representations of context due to
being ill-defined, discrete and incomparable, and not reflective of the 
diversity of human language \cite{ruder2016towards}. In context aware language
models, the notion of a domain is replaced with a set of context variables that
each describe some aspect of the associated language such as the topic, time, 
or language. These variables can be dynamically combined to create a continuous representation of context as a low-dimensional embedding \cite{TangContextAware}.
The context variables and context embedding can then be used to adapt a recurrent
neural network language model (RNNLM).

The standard approach for using a context embedding to adapt an RNNLM is to simply
concatenate the context representation with the word embedding at the input to the
RNN \cite{mikolov2012context}. Optionally, the context embedding is also concatenated
with the output from the recurrent layer so that the output layer can be adapted 
as well. This basic strategy has been adopted for various types of adaptation such
as for LM personalization \cite{wen2013recurrent,li2016persona}, adapting an LM to different genres of television shows \cite{chen2015recurrent}, adapting to long range dependencies in a document \cite{Ji2015DocumentCL}, sharing information in generative text classifiers \cite{Yogatama2017GenerativeAD}, and in other cases as well.

In this paper, we study methods of improving the mechanism for using context variables for adapting an
RNNLM. The standard approach of adapting the hidden layer is equivalent to an additive
transformation of the hidden state. We propose complimenting this with a multiplicative rescaling at the 
hidden layer and show that it consistently helps when the language model is used
as a generative text classifier and can sometimes improve perplexity. 

Using context dependent bias vectors is one way to adapt the output layer but it becomes infeasible when both the vocabulary size and the number of contexts are large. The method from \newcite{mikolov2012context} of using the low-dimensional context embedding to adapt the output layer avoids the excessive memory issue of context-dependent bias vectors but our experiments show that it does not capture isolated but important  details.
%%MO: I don't like "inadequate", you  could also say "it is limited in its ability to .."
%is inadequate both in terms of perplexity and classification metrics. 
We propose a hashing technique to simultaneously benefit from context-dependent weights and avoid the high memory cost. The combination of the low-rank and hashing techniques for adapting the output layer shows a consistent improvement across our experiments on three different corpora.

\section{Model}

Our model is built on top of a standard RNN language model. There are three key parts which
we discuss below: how we represent context using a low-dimensional
embedding, the mechanism for using the context embedding for adapting the recurrent 
layer, and the mechanisms for adapting the output layer.

\subsection{Representing outside context}

We assume access to one or more indicator variables, $c_{1:n}=c_1, c_2, \dots c_n$, that hold
information about the outside context for each sentence. These can be indicators for
topic, geographic region, time period, or other meta-data. In \cite{mikolov2012context}
LDA topic vectors are used for the outside context. In \cite{TangContextAware} the outside
context is a sentiment score and a product id for a product review dataset. We adopt their
method of combining information from multiple context variables using a simple neural
network. This strategy is well-suited for the types of context variables that we will
see in our experiments, such as speaker identity. In other cases, it may be more appropriate
to use topic models \cite{chen2015recurrent,ghosh2016contextual} or an RNN
\cite{Hoang2016IncorporatingSI} to build the context representation.

For each context variable $c_i$,
%$c_1, c_2, \dots c_n$, 
we learn an associated embedding matrix $\mathbf{E}_{i}$, $i=1,\ldots, n$.
%$\mathbf{E}_{1}, \mathbf{E}_{2}, \dots \mathbf{E}_{n}$. 
If $n=1$ then the embedding
can directly be used as the context representation. Otherwise, a single layer neural network is used to combine the embeddings from the individual variables.
\[ \vec{c} = tanh( \sum_{i} \mathbf{M}_i \mathbf{E}_i c_i + b_0)\]
$\mathbf{M}_i$ and $b_0$ are parameters learned by the model. The context embedding, $\vec{c}$, is used for adapting both the hidden and the output layer of the RNN.

\subsection{Adapting the hidden layer}

The equation for the hidden layer of an RNN is 
\[ s_t = \sigma ( \mathbf{U} \vec{w}_t + \mathbf{S} s_{t-1} + b_1) \]
where $\vec{w}_t$ is the word embedding of the $t$-th word, $s_{t-1}$ is the hidden state from the previous time step and $\sigma$ is the activation function. To make use of the context embedding, $\vec{c}$, for adapting the hidden layer the term $\mathbf{F} \vec{c}$ is inserted resulting in
\[s_t = \sigma ( \mathbf{U} \vec{w}_t + \mathbf{S} s_{t-1} + \mathbf{F} \vec{c} + b_1)\]
We refer to the insertion of the $\mathbf{F} \vec{c}$ term as an additive adaptation of the hidden layer. It is equivalent to the unadapted version except with an adapted bias term. It can be implemented by simply concatenating the context vector $\vec{c}$ with the word embedding $\vec{w}_t$ at each timestep at the input to the recurrent layer.

To increase the adaptability of the hidden layer we use a context-dependent multiplicative rescaling of the hidden layer weights. The method is borrowed from \newcite{ha2016hypernetworks} where it is used for dynamically adjusting the parameters of a language model in response to the previous words in the sentence.
Using this row rescaling technique on top of the additive adaptation from above, the equation becomes
\[ s_t = \sigma ( \mathbf{C}_u \vec{c} \odot \mathbf{U} \vec{w}_t + \mathbf{C}_w \vec{c} \odot \mathbf{S} s_{t-1} + 
\mathbf{F} \vec{c} + b_1) \]
where $\mathbf{C}_u$ and $\mathbf{C}_w$ are parameters of the model and $\odot$ is the elementwise 
multiplication operator. The element-wise multiplication is a low-cost operation and can even be 
pre-calculated so that model evaluation can happen with no extra computation compared to a vanilla RNN.

\subsection{Adapting the output layer}

The output probabilities of an RNN are given by $y_t = softmax(\mathbf{V}s_t + b_2)$. In our case, 
we tie the weights between the word embeddings in the input and output layer: $\mathbf{W}^T = \mathbf{V}$ 
\cite{press2016using,inan2016tying}.

One way of adapting the output layer is to let each context have its own bias vector. This requires the
use of a matrix of size $|V| \times |C|$, which may be intractable when both $|V|$ and $|C|$ are large.
Here, $|V|$ is the size of the vocabulary and $|C|$ is the total number of possible contexts.
\newcite{mikolov2012context} use a low-rank factorization of of the adaptation matrix, replacing the $|V| \times |C|$ matrix with the product of a matrix $\mathbf{G}$ of size $|V| \times k$ and a context embedding $\vec{c}$ of size $k$. 
\[ y_t = softmax(\mathbf{V} s_t + \mathbf{G} \vec{c} + b_2) \]
The total number of parameters is now a much more manageable $O(|V| + \sum_i |C_i|)$ instead of $O(\sum_i |V| |C_i|)$.
The advantage of a low-rank adaptation is that it forces the model to share information between similar contexts. The disadvantage, is that important differences between similar contexts can be lost.

We employ feature hashing to reduce the memory requirements but retain some of the benefits of having an individual bias term for each context-word pair. The context-word pairs are hashed into buckets and individual bias terms are learned for each bucket. The hashing technique relies on having direct access to the context variables $c_{1:n}$. Representing context as a latent topic distribution precludes the use of this hashing adaptation.

The choice of hashing function is motivated by what is easy and fast to perform inside 
the Tensorflow computation graph framework. If $w$ is a word id and $c_{1_n}$ are context variable ids then the hash table index is computed as  
\[ h_i(w, c_i) = w r_0 + c_i r_i \bmod l\]
where $l$ is the size of the hash table and $r_0$ and the $r_i$'s are all
fixed random integers. The value of $l$ is usually set to a large prime
number. The function $H : \mathbb{Z} \rightarrow \mathbb{R}$ maps hash indices
to hash values and is implemented as a simple array.

Since $l$ is much smaller than the total number of inputs, there will be many 
hash collusions. Hash collusions are known to negatively effect the perplexity
\cite{mikolov2011strategies}. To deal with this issue, we restrict the hash table
to context-word pairs that are observed in the training data. A Bloom filter data
structure records which context-word pairs are eligible to have entries in the
hash table. The design of this data structure trades off a compact representation 
of set membership against a small probability of false positives 
\cite{bloom1970space,talbot2008randomized,xu2011randomized}. A small amount of
false positives is relatively harmless in this application because they do not
impair the ability of the Bloom filter to eliminate almost all of the hash collusions.

The function $\beta : \mathbb{Z} \rightarrow [0, 1]$ is used by the Bloom filter to
map hash indices to binary values.
\[ B(w, c_i) = \prod_{j=1}^{16} \beta(h_{i,j}(w, c_i)) \]
The hash functions $h_{i,j}$ are defined in the same way as the $h_i$'s above
except that they use distinct random integers and the size of the table, $l$,
can be different. Because $\beta$ is a binary function, the product $B(w, c_i)$ will
always be zero or one. Thus, any word-context pairs not found in the Bloom filter
will have their hash values set to zero.

The final expression for the hashed adaptation term is given by
\[ Hash(w, c_{1:n}) = \sum_{i=1}^n H(h_i(w, c_i)) B(w, c_i) \]
\[ y_t = softmax(\mathbf{V} s_t + \mathbf{G} \vec{c} + b_2 + Hash(w_t, c_{1:n}))\]

\section{Data}

The experiments make use of three corpora chosen to give a diverse prospective on adaptation in language modeling. Summary information on the datasets (Reddit, Twitter, and SCOTUS) is provided in Table \ref{table:data}
and each source is discussed individually below. The Reddit and SCOTUS data are tokenized and lower-cased 
using the standard NLTK tokenizer \cite{BirdKleinLoper09}. 

\begin{table}[]
\centering
\begin{tabular}{crrl}
\textbf{Source} & \multicolumn{1}{c}{\textbf{Size}} & \multicolumn{1}{c}{\textbf{Vocab.}} & \multicolumn{1}{c}{\textbf{Context}} \\ \hline
Reddit & 8,000K & 68,000 & Subreddit \\
Twitter & 77K & 194 & Language \\
SCOTUS & 864K & 18,000 & Case, Spkr., Role 
\end{tabular}
\caption{Number of sentences, vocabulary size and context variables for the three corpora.}
\label{table:data}
\end{table}

\paragraph{Reddit} Reddit is the world's largest online discussion forum and is comprised of 
thousands of active subcommunities dedicated to a wide variety of themes. % \cite{jaech2015talking}
Our training data is 8 million sentences from Reddit comments during the month of April 2015. The 68,000 word vocabulary is selected by taking all tokens that occur at least 20 times in the training data. The remaining tokens are mapped to a special UNK token leaving us with an OOV rate of 2.3\%.

The context variable is the identity of the subreddit, i.e. community, that the comment came from.
There are 5,800 subreddits with at least 50 training sentences. The remaining ones are grouped
together in an UNK category. The largest subreddit occupies just 4.5\% of the data and the perplexity
of the subreddit distribution is 742. By using a large number of subreddits, we highlight an advantage 
of model adaptation which is to be able to use a single unified model instead of training thousands of separate models for each individual community. Similarly, using context dependent bias vectors for this data instead of the hash adaptation would require learning 400 million additional parameters.

\paragraph{Twitter} 
The Twitter training data has 77,000 Tweets each annotated with one of nine languages: English, German,
Italian, Spanish, Portuguese, Basque, Catalan, Galician, and French. The corpus was collected by combining
resources from published data for language identification tasks during the past few years. Sentences
labeled as unknown, ambiguous, or containing code-switching were not included. The data is 
unbalanced across languages with more than 32\% of the Tweets being Spanish and the smallest four 
languages (Italian, German, Basque, and Galician) all individually less than 1.5\% of the total. 
There are 194 unique character tokens in the vocabulary. Graphemes that are surrogate-pairs in the UTF-16 encoding, such 
as emoji, are split into multiple vocabulary tokens. No preprocessing or tokenization is performed on this
data except that newlines were replaced with spaces for convenience.

\paragraph{SCOTUS} Approximately 864,000 sentences of training data spanning arguments from 1990-2011.
These are speech transcripts from the United States Supreme Court. Utterances are labeled with the 
case being argued (n=1,765), the speaker id (n=2,276), 
% perplexity of cases is 1737 and perplexity of speakers is 294
and the speaker role (justice, advocate, or unidentified). These three context variables are defined
in the same way as in \newcite{hutchinson2013exceptions}, where a small portion of this data was used
in language modeling experiments. The vocabulary size is around 18,000 words. Utterances longer than
45 words were split into smaller utterances. 

\section{Experiments}

In these experiments we fix the size of the word embedding dimensions and recurrent layers so 
as not to exhaust our computational resources and then vary the different mechanisms for adapting the model. We used an LSTM with coupled input and forget gates for a 20\% reduction in computation 
time \cite{greff2016lstm}. Dropout was used as a regularizer on
the input and outputs of the recurrent layer as described in \newcite{zaremba2014recurrent}.
When the vocabulary is large, computing the full cross-entropy loss can be prohibitively
expensive. For the large vocabulary experiments, we used a sampled softmax strategy with a unigram distribution to speed up training \cite{Jean2015OnUV}.

A summary of the key hyperparameters for each class of experiments if given in 
Table \ref{table:hyperparams}. The total parameter column in this table is based on the 
unadapted model. Adapted models will have more parameters depending on the type of adaptation. When using hash adaptation of the output layer, the size of the Bloom filter is 100 million and the size of the hash table is 80 million. The model is implemented using the Tensorflow library.\footnote{See \url{https://github.com/ajaech/calm} for code.} Optimization is done using Adam with a learning rate of 0.001. Each model trained in under three days using 8 CPU threads.

\begin{table}[]
\centering
\begin{tabular}{cccc}
\textbf{Parameter} & \textbf{Reddit} & \textbf{SCOTUS} & \textbf{Twitter} \\ \hline
Batch Size & 400 & 300 & 200 \\
Word Embed. & 200 & 200 & 30 \\
LSTM Size & 240 & 240 & 200 \\
Dropout & 0\% & 15\% & 10\% \\
Neg. Samples & 100 & 100 & NA \\
Total Params & 14M & 4M & 300K
\end{tabular}
\caption{Summary of Key Hyperparamters}
\label{table:hyperparams}
\end{table}

Although the model is trained as a language model, it can be used as a generative text classifier.
%The classification rule is given by $argmax_{c_i} \sum_j \log p(w_j| w_{1:j-1}, c_{k \neq i}, c_i)$.
When there are multiple context variables, we treat all but one of them as known values and attempt to identify the unknown one. It is not necessary to compute the probabilities over the full vocabulary. The sampled softmax criteria can be used to greatly speed up evaluation of the classifier. 

\subsection{Reddit Experiments}

The size of the subreddit embeddings was set to 25. Table \ref{table:reddit} gives the perplexities and average AUCs for subreddit detection for different adapted models.%\footnote{Additional experiment results are included in the appendix.}
 The evaluation data contains 60,000 sentences. For comparison, an unadapted 4-gram Kneser-Ney model trained on the same data has a perplexity of 119. The models with the best perplexity do not use multiplicative adaptation of the hidden layer, but it is useful in the detection experiments. 

\begin{table}[]
\centering
\begin{tabular}{cc|cc|cr|c}
\multicolumn{2}{c}{\textbf{Hidd.}} & \multicolumn{2}{c}{\textbf{Output}} & \multicolumn{1}{l}{} & \multicolumn{1}{l}{} & \multicolumn{1}{l}{} \\
\textbf{$\times$} & \textbf{$+$} & \multicolumn{1}{l}{\textbf{LR}} & \multicolumn{1}{l}{\textbf{Hash}} & \textbf{PPL} & \multicolumn{1}{l}{$\Delta$\textbf{PPL}} & \textbf{AUC} \\ \hline
N & N & N & N & 75.2 & -- & -- \\
N & N & N & Y & 69.6 & 7.3\% & 76.5 \\
N & N & Y & N & 68.0 & 9.5\% & 75.5 \\
N & Y & N & Y & 66.9 & 11.0\% & 78.9 \\
N & Y & Y & N & 68.0 & 9.6\% & 75.3 \\
N & Y & Y & Y & \textbf{66.5} & \textbf{11.5\%} & 78.4 \\
Y & N & Y & Y & 67.2 & 10.6\% & 78.9 \\
Y & Y & Y & N & 68.3 & 9.1\% & 75.7 \\
Y & Y & Y & Y & 67.1 & 10.7\% & \textbf{79.2} \\
\end{tabular}
\caption{Perplexities and Classification Avg. AUCs for Reddit Models}
\label{table:reddit}
\end{table}

We can inspect the context embeddings learned by the model to see if it is exploiting
similarities between subreddits in the way that we expect. Table \ref{table:neighbors}
lists the nearest neighbors by Euclidean distance to three selected subreddits. We can
see that the nearest neighbors match our intuitions. The closest subreddits to Pittsburgh
are communities created for other big cities and states. The Python subreddit is close to
other programming languages' communities, and the NBA subreddit is close to the communities
for individual NBA teams.

\begin{table}[]
\centering
\begin{tabular}{ccc}
\textbf{Pittsburgh} & \textbf{Python} & \textbf{NBA} \\ \hline
Atlanta & CSharp & Warriors \\
Montana & JavaScript & Rockets \\
MadisonWI & CPP\_Questions & Mavericks \\
Baltimore & CPP & NBASpurs
\end{tabular}
\caption{Nearest neighbors to selected subreddits in the context embedding space.}
\label{table:neighbors}
\end{table}

The subreddit detection involves predicting the subreddit a given comment came from with eight subreddits to choose from (AskMen, AskScience, AskWomen, Atheism, ChangeMyView, Fitness, Politics, and Worldnews) and nine distractors (Books, Chicago, NYC, Seattle, ExplainLikeImFive, Science, Running, NFL, and TodayILearned).\footnote{These are the same subreddit used in \newcite{tran2016characterizing} for a related but not comparable classification task.} To make a classification decision we evaluate the perplexity of each
comment under the assumption that it belongs to each of the eight subreddits. We use z-score normalization across the eight perplexities to create a score for each class. The predictions are evaluated by averaging the AUC of the eight individual ROC curves. The best model for the classification task uses all four types of adaptation. Interestingly, the multiplicative
adaptation of the hidden layer is clearly useful for classification even though it does not help with perplexity.

The perplexities for selected large subreddits are listed in Table \ref{table:ppl_comparison}. 
It can be seen that the relative gain from adaptation is largest when the topic of the subreddit is more narrowly
focused. The biggest gains were achieved for subreddits dedicated to specific sports, tv shows, or video games.
Whereas, the gains were smallest for subreddits like Videos or Funny whose content tends to be more
diverse. The knowledge that a sentence came from a pro-wrestling subreddit effectively provides more information
about the text than the analogous piece of knowledge for the Pics or Videos subreddit. This would seem to indicate
that further gains could be possible if additional contextual information could be provided. An alternative explanation,
that subreddits with less sentences in the training data receive more benefit from adaptation, is not supported by
the data.

\begin{table*}[]
\centering
\begin{tabular}{lrrrl}
\multicolumn{1}{c}{\textbf{Subreddit}} & \multicolumn{1}{c}{\textbf{Base. PPL}} & \multicolumn{1}{c}{\textbf{Adapt. PPL}} & \multicolumn{1}{c}{$\Delta$\textbf{PPL}} & \multicolumn{1}{c}{\textbf{Description}} \\ \hline
FlashTV & 90.5 & 68.2 & 24.6\% & A popular TV show \\
shield & 99.4 & 77.3 & 22.2\% & A tv show \\
GlobalOffensive & 97.1 & 79.3 & 18.3\% & A PC video game \\
nba & 103.3 & 86.4 & 16.3\% & National Basketball Association \\
SquaredCircle & 85.7 & 71.7 & 16.3\% & Professional Wrestling \\
Fitness & 50.1 & 42.3 & 15.5\% & Exercise and fitness \\
hockey & 85.5 & 72.4 & 15.2\% & Professional hockey \\
leagueoflegends & 71.1 & 61.0 & 14.3\% & A PC video game \\
pcmasterrace & 71.7 & 62.0 & 13.5\% & PC gaming \\
nfl & 84.2 & 74.0 & 12.2\% & National Football League \\
AskWomen & 62.1 & 55.3 & 10.9\% & Questions for women \\
news & 70.8 & 65.0 & 8.2\% & General news stories and discussion \\
worldnews & 85.7 & 79.7 & 7.1\% & Global news discussion \\
AskMen & 69.4 & 66.7 & 3.9\% & Questions for men \\
gaming & 79.0 & 76.1 & 3.7\% & General video games interest group \\
pics & 74.0 & 71.8 & 3.0\% & Funny or interesting pictures \\
videos & 62.9 & 61.1 & 2.9\% & Funny or interesting videos \\
funny & 72.6 & 70.8 & 2.5\% & Sharing humorous content \\
\end{tabular}
\caption{Comparison of perplexities per subreddit}
\label{table:ppl_comparison}
\end{table*}

%\begin{figure}[h]
%\centering
%\includegraphics[width=0.5\textwidth]{logodds}
%\caption{Histogram of log odd ratio between true model and adapted model.}
%\label{fig:logodds}
%\end{figure}

%Figure \ref{fig:logodds} shows a histogram of the log odds ratio for each word
%between the distribution as calculated on the training data and the same distribution
%as calculated by an adapted model: $\log_{10} \frac{P_\mathrm{true}(w)}{P_\mathrm{adapted}(w)}$.
%Most words have their log odds ratio near zero, meaning that the low-rank adapted model well-approximates
%their true probabilities. There are a small number of words on either tail, however, that are not well-approximated
%by the low-rank adaptation. Example words from the Seattle subreddit on the right tail are names of cities or 
%neighborhoods in the region such as ``Bellevue'' and ``Seattle''. An example word from the left tail is ``Toronto''.
%The words ``Toronto'' and ``Seattle'' behave similarly in almost every context except in the Seattle Subreddit where
%they behave very differently.

\subsection{Twitter experiments}

The Twitter evaluation was done on a set of 14,960 Tweets. The language context embedding vector 
dimensionality was set to 8. When both the vocabulary and the number of contexts are small,
as in this case, there is no danger of hash collusions. We disable the bloom filter making
the hash adaptation essentially equivalent to having context dependent bias vectors.

\begin{table}[]
\centering
\begin{tabular}{cc|cc|ccl}
\multicolumn{2}{c}{\textbf{Hidden}} & \multicolumn{2}{c}{\textbf{Output}} & \multicolumn{1}{l}{} & \multicolumn{1}{l}{} &  \\
$\times$ & $+$ & \multicolumn{1}{l}{\textbf{LR}} & \multicolumn{1}{l}{\textbf{Hash}} & \textbf{PPL} & \textbf{Acc.} & \textbf{F1} \\ \hline
N & N & N & N & 6.44 & -- &  -- \\
N & N & N & Y & 6.43 & 56.1 & 44.0 \\
N & N & Y & N & 6.37 & 49.7 & 36.6 \\
N & Y & Y & N & 6.21 & 91.4 & 82.9 \\
N & Y & N & Y & 6.25 & 92.5 & 84.4 \\
N & Y & Y & Y & \textbf{6.15} & 92.8 & 85.2 \\
Y & N & Y & N & 6.28 & 93.2 & 85.1 \\
Y & Y & Y & N & 6.54 & \textbf{94.2} & \textbf{86.3} \\
Y & Y & Y & Y & 6.35 & 93.3 & 85.9 \\
\end{tabular}
\caption{Results on Twitter data.}
\label{table:twitter}
\end{table}

Table \ref{table:twitter} reports the results of the experiments on the Twitter corpus. We compute
both the perplexity and measure the performance of the models on a language identification task.
In terms of perplexity, the best models do not make use of the multiplicative hidden layer adaptation,
consistent with the results from the Reddit corpus. In general, the improvement in perplexity from
adaptation is small (less than 5\%) on this corpus compared to our other experiments where we saw
relative improvements two to four times as big. This is likely because the LSTM can figure out by 
itself which language it is modeling early on in the sequence and adjust its predictions accordingly. 

\begin{figure}[h]
\centering
\includegraphics[width=0.5\textwidth]{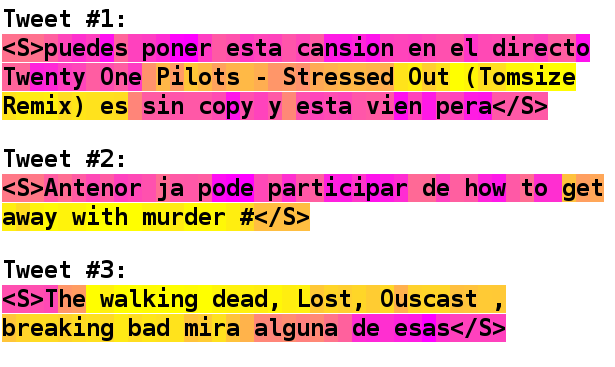}
\caption{The value of the dimension of the hidden vector that is most correlated with Spanish text for
three different code-switched Tweets.}
\label{fig:heatmap}
\end{figure}

To investigate this further, we trained a logistic regression classifier to predict the language using the
state from the LSTM at the last time step on the unadapted model as a feature vector. 
% got an accuracy of 92.1\% and an F1 score of 77.9. 
Using just 30 labeled examples per class it is possible to get 74.6\% accuracy and a 49.3 F1 score. 
Furthermore, we find that a single dimension in the hidden state of the unadapted model is often 
enough to distinguish between different languages even though the model was not given any supervision
signal \cite{karpathy2015visualizing,radford2017}. Figure \ref{fig:heatmap} visualizes the value of the 
dimension of the hidden layer that is the strongest indicator of Spanish on three different
code-swtiched tweets. Code-switching is not a part of the training data for the model but
it provides a compelling visualization of the ability of the unsupervised model to quickly
recognize the language. The fact that it is so easy for the unadapted model to pick-up on the
identity of the contextual variable fits with our explanation for the small relative gain
in perplexity from the adapted models.

Our best model, using multiplicative adaptation of the hidden layer, achieves an accuracy of 94.2\% on this task. That is a 19\% relative reduction in the error rate from the best model without multiplicative adaptation.

\subsection{SCOTUS experiments}

Table \ref{table:scotus} lists the results for the experiments on the SCOTUS corpus. The size of
the context embeddings are 9, 15, and 8 for the case, speaker, and role variables respectively.
For calculating perplexity we use 60,000 sentence evaluation set. For the classification experiment we selected 4,000 sentences from the test data from eleven different justices and attempted to classify the identity of the justice. The perplexity of the distribution of judges over those sentences is 8.9 (11.0 would be uniform). So, the data is roughly balanced. When classifying justices, the model is given the case context variable, but we do not make any special effort to filter candidates based on who was serving on the court during that time, i.e. all eleven justices are considered for every case. 

\begin{table}[]
\centering
\begin{tabular}{cc|cc|cc|c}
\multicolumn{2}{c}{\textbf{Hidden}} & \multicolumn{2}{c}{\textbf{Output}} & \textbf{} & \textbf{} \\
$\times$ & $+$ & \textbf{LR} & \textbf{Hash} & \textbf{PPL} & $\Delta$\textbf{PPL} & \textbf{ACC} \\ \hline
N & N & N & N & 37.3 & -- & -- \\
N & N & N & Y & 31.2 & 16.5\% & 29.6 \\
N & N & Y & N & 32.9 & 12.0\% & 26.2 \\
N & Y & Y & N & 32.7 & 12.4\% & 25.4 \\
N & Y & Y & Y & 29.8 & 20.3\% & 31.1 \\
Y & N & Y & N & 32.3 & 13.4\% & 24.5 \\
Y & Y & Y & N & 32.2 & 13.7\% & 26.1 \\
Y & N & Y & Y & \textbf{29.2} & \textbf{21.7\%} & \textbf{32.4} \\
Y & Y & Y & Y & 29.4 & 21.1\% & 31.9 \\
\end{tabular}
\caption{Results on the SCOTUS data in terms of perplexity and classification accuracy (ACC) for the justice identification task.}
\label{table:scotus}
\end{table}

For both the perplexity and classification metrics, the hash adaptation makes a big difference.
The model that uses only hash adaptation and no hidden layer adaptation has a better perplexity
than any of the model variants that use both hidden adaptation and low-rank adaptation of the output layer. 

\begin{table}[]
\centering
\begin{tabular}{cccc}
\textbf{Case} & \textbf{Spkr.} & \textbf{Role} & \textbf{PPL} \\ \hline
N & N & N & 37.3 \\
N & N & Y & 36.5 \\
N & Y & N & 33.6 \\
N & Y & Y & 33.3 \\
Y & N & N & 31.5 \\
Y & N & Y & 30.3 \\
Y & Y & N & 29.6 \\
Y & Y & Y & 29.4 
\end{tabular}
\caption{Perplexities for different combinations of context variables on the SCOTUS corpus.}
\label{table:scotus_context}
\end{table}

To ascertain which of the context variables have the most impact, we trained additional models
with using different combinations of context variables. The model architecture is the one that
uses all four forms of adaptation. Results are listed in Table \ref{table:scotus_context}. The
most useful variable is the indicator for the case. The role variable is highly redundant---
almost every speaker only appears in a single role. The experiments indicate that the role
variable provides useful information to the model, and the knowledge of the speaker 
identity seems to not convey much useful information beyond what is provided by the role.

\begin{table*}[]
\centering
\begin{tabular}{ccl}
\textbf{Spkr.} & \textbf{Role} & \textbf{Sentence} \\ \hline
Roberts & J. & \texttt{We'll hear argument first this morning in Ayers.} \\
Breyer & J. & \texttt{I mean, I don't think that's right.} \\
Kagan & J. & \texttt{Well, I don't think that's right.} \\
Kagan & A. & \texttt{Mr. Chief Justice, and may it please the court:} \\
Bork & A. & \texttt{--No, I don't think so, your honor.} \\
\end{tabular}
\caption{Sentences generated from the adapted model using beam search under different
 assumptions for speaker and role contexts.}
\label{table:scotus_examples}
\end{table*}

%\begin{table*}[]
%\centering
%\begin{tabular}{cccp{9cm}}
%\textbf{Speaker} & \textbf{Role} & \textbf{Case} & \textbf{Sentence} \\ \hline
%Scalia & J. & 1 & \texttt{Assuming what is obscene in this particular category that's we assume?} \\
%Morazzini & A. & 1 & \texttt{As the court found it indicated they did not use the hypothetical as a -- when the video records but minors are minors in any manner.} \\
%Morazzini & A. & 2 & \texttt{We believe that that 's a intrusion on the general rights of the united states which suggests guns are protected}
%\end{tabular}
%\caption{Sentences selected from the adapted model using greedy decoding under different
% assumptions for Speaker, role, and case contexts.}
%\label{table:scotus_examples2}
%\end{table*}

In Table \ref{table:scotus_examples} we list sentences generated from the fully adapted model
(same one as the last line in Table \ref{table:scotus}) using beam search. The value of the
context variable for the Case is held fixed while we explore different values for the 
Speaker and Role variables. Anecdotally, we see that the model captures some information
about John Roberts role as chief justice. The model learns that Justice Breyer tends to
start his questions with the phrase ``I mean'' while Justice Kagan tends to start with
``Well''. Roberts and Kagan appear in our data both as justices and earlier as advocates.

%Table \ref{table:scotus_examples2} is similar except the sentences are generated using
%random sampling rather than beam search. Here we see the ability of the model to make
%use of words relating to the topic of the case being argued. The first case is about
%if violent video games are protected under the 1st amendment. The generated sentences
%include the relevant words ``obscene'', ``video'', and ``minors''. The second sentence in the
%table is generated using Mr. Morazzini as the speaker, the lawyer who argued this case.
%The third sentence uses the same speaker context but a different case this time about
%the 2nd amendment. Even though Mr. Morazzini never talks about guns or the 2nd amendment
%in the training data, the model is capable of generating a sentence under the counter-factual
%assumption that he argued the case about gun rights.

\section{Related Work}

Multiple survey papers cover the early history of language model adaptation
\cite{demori1999language,bellegarda2004statistical}. We mention just the most
recent closely related work here.

%Our experiments draw heavily on the ideas from \newcite{mikolov2012context}.

The multiplicative rescaling of the recurrent layer weights is used in the Hypernetwork model \cite{ha2016hypernetworks}. The focus of this model is to allow the LSTM to adjust automatically depending on the context of the previous words. This is different from our work in that we are
adapting based on contextual information external to the word sequence. \newcite{gangireddy2016unsupervised}
also use a rescaling of the hidden layer for adaptation but it is done as a fine-tuning step and not during
training like our model.

The RNNME model from \newcite{mikolov2011strategies} uses feature hashing to train a maximum entropy model
alongside an RNN language model. The setup is similar to our method of using hashing to learn context-dependent
biases. However, there are a number of differences. The motivation for the RNNME model was to speed-up training
of the RNN not to compensate for the inadequacy of low-rank output layer adaptation, which had yet to be 
invented. Furthermore, \newcite{mikolov2011strategies} do not use context dependent features in the max-ent component
of the RNNME model nor do they have a method for dealing with hash collusions such as our use of Bloom filters.

The idea of having one part of a language model be low-rank and another part to be an additive correction
to the low-rank model has been investigated in other work \cite{eisenstein2011sparse,hutchinson2013exceptions}.
In both of these cases, the correction term is encouraged to be sparse by including an L1 penalty. Our implementation
did not promote sparsity in the hash adaptation features but this idea is worth further consideration. The hybrid LSTM and count based language model is an alternative way of correcting for a low-rank approximation \cite{neubig2016generalizing}.

\newcite{Hoang2016IncorporatingSI} studies how to incorporate side information into an RNN 
language model. For their data, they claim a bigger win by adapting at the output layer 
rather than the hidden layer. (This matches our own observations on the Reddit and SCOTUS
data.) Their work did not address adapting at both the hidden and output layers simultaneously.
Most work on adaptation does not consider combining multiple context factors but there are 
some exceptions \cite{hutchinson2013exceptions,TangContextAware,Hoang2016IncorporatingSI}.

%\newcite{Ragni2016MultiLanguageNN} do multi-lingual language modeling. This is the same
%task as the Twitter LM that we did except they did it at the word level and they use
%shared but unadapted recurrent layer weights.

\section{Conclusions \& Future Work}

While our results suggest that there is not a one-size-fits-all approach to
language model adaptation, it is clear that we improve over the standard adaptation 
approach. The model from \newcite{mikolov2012context}, equivalent to using just
additive adaptation on the hidden layer and low-rank adaptation of the output layer, is outperformed for all three datasets at both the language modeling and classification tasks. For language modeling, the multiplicative hidden layer adaptation was only helpful for the SCTOUS dataset. However, the combined low-rank and hash adaptation of the output layer consistently gave the best perplexity. For the classification tasks, the multiplicative hidden layer adaptation is clearly useful, as is the combined low-rank and hash adaptation of the output layer.

Importantly, there is not always a strong relationship between perplexity and classification 
scores. Our results may have implications for work on text generation where it can be more desirable
to have more control over the generation rather than the lowest perplexity model. 
More studies are needed to get intuition about what types of context variables will provide the most
benefit. Our investigation of the language context in the Twitter experiments gives a useful
takeaway: context variables that are easily predictable from the text alone are unlikely to be helpful.

In future work, we would like to consider additional mechanisms for using the context embedding $\vec{c}$
to adapt the LSTM parameters. We also plan to extend our hash adaptation to incorporate longer word histories, rather than just unigrams combined with context.

\bibliography{mybib}
\bibliographystyle{emnlp_natbib}

\end{document}